\documentclass{article}
\usepackage{spconf,amsmath,graphicx}
\usepackage{hyperref}
\usepackage{multicol}
\usepackage{subcaption}
\usepackage{amsfonts}
\usepackage{subcaption}
\usepackage{float}

\title{Transformer  based  Reinforcement  Learning for Games}
\name{Uddeshya Upadhyay$^{\dagger}$, Nikunj Shah$^{\star}$ Sucheta Ravikanti$^{\alpha}$, Mayanka Medhe$^{\dagger}$} 
\address{
    $^{\dagger}$Department of Computer Science and Engineering, Indian Institute of Technology-Bombay\\
    $^{\star}$Department of Mechanical Engineering, Indian Institute of Technology-Bombay\\
    $^{\alpha}$Department of Electrical Engineering, Indian Institute of Technology-Bombay
}

\begin{document}
\maketitle

\nocite{a1, a2, a3, a5, a6, a7, a8, a9, a10, a11}
\begin{abstract}
Recent times have witnessed sharp improvements in reinforcement learning tasks using deep reinforcement learning
techniques like Deep Q Networks, Policy Gradients, Actor
Critic methods which are based on deep learning based models and back-propagation of gradients to train such models.
An active area of research in reinforcement learning is about
training agents to play complex video games, which so far has
been something accomplished only by human intelligence.
Some state of the art performances in video game playing
using deep reinforcement learning are obtained by processing
the sequence of frames from video games, passing them through
a convolutional network to obtain features and then using
recurrent neural networks to figure out the action leading to
optimal rewards. The recurrent neural network will learn to
extract the meaningful signal out of the sequence of such features. In this work, we propose a method utilizing transformer networks which have recently replaced RNNs in Natural Language Processing (NLP), and perform experiments to
compare with existing methods.
\end{abstract}
\section{Introduction and Related Work}
\label{sec:intro}
Recent advancements in reinforcement learning have witnessed the heavy use of Deep Neural Networks (DNN) to perform many of the reinforcement learning tasks such as prediction and control. These classes of approaches at the intersection of deep learning and reinforcement learning are known as Deep reinforcement learning (DRL). DRL uses deep learning and reinforcement learning principles to create efficient algorithms that can be applied on areas like robotics, video games, finance, and healthcare \cite{franccois2018introduction}. Implementing
neural networks like deep convolutional networks with reinforcement learning algorithms such as Q-learning, Actor
Critic or Policy Search results in a powerful model (DRL)
that is capable to scale to previously unsolvable problems \cite{mnih2013playing}. That is because DRL usually work with raw sensors or image signals as input (for instance in  Deep Q Networks - DQN for ATARI games \cite{Arulkumaran_2017}) and can receive the benefit of end-to-end reinforcement learning as well as that of convolutional neural networks. In other words, neural networks act as function approximators, of action-value functions used in predicting the next best action, which is particularly useful in reinforcement learning when the state or action space is too large to be completely known or to be stored in memory.

Recently, DQN with recurrent layers also called as
Deep Recurrent Q Networks (DRQN), have started showing promising results in reinforcement learning problem. They have the capability to outperform the DQN’s outcomes and generate better trained agents in certain scenarios. The primary reason for this being, DQN has limited or no amount of distant history. In practice, DQNs are often trained with just a single state representation corresponding to current time-steps (i.e often times they neglect the temporal aspects of the input), even when fed with the sequence of state representation, standard DQN architecture without RNNs are unable to capture and extract from the temporal aspect of the sequences. Thus DQN will be unable to master games that require the player to remember events more distant in the past. Put differently, any game that requires a memory of past states in the trajectory will appear non-Markovian because the future game states (and rewards) depend on more than just DQN’s current input. Instead of a Markov Decision Process (MDP), the game becomes a Partially-Observable Markov Decision Process (POMDP) \cite{hausknecht2015deep}. Hence, LSTM along with DQN is used in such cases.

Since LSTM or recurrent neural networks, in general, have had a strong presence in the Natural Language Processing (NLP), we tried to implement a technique inspired from NLP called Transformer to perform the reinforcement learning tasks.
Transformer was first introduced in \cite{vaswani2017attention} to perform sequence-to-sequence translation, however, it has been adapted successfully in various applications spanning language, speech, etc.

NLP uses a sequence-to-sequence architecture at the core in various tasks. In NLP we generally need to analyze a sequence of words and generate another sequence of words based on them, therefore language translation task is one such example where sequence to sequence architecture is applicable. LSTM was a popular choice to be used as an encoder and decoder for the above-specified task and related architecture. LSTM based approach implicitly accounts for 'attention' which is a mechanism that looks at an input sequence and decides at each step which other parts of the sequence are important. The transformer was a novel technique introduced to replace this attention-mechanism performed by LSTM by more effective and explicit attention-mechanism. It is also an encoder-decoder model but differs from LSTM by avoiding any usage of recurrent neural networks which is common in LSTM, GRU, etc. This improves training time as well as accuracy in NLP tasks.

\section{Methods and Experiments}
\label{sec:method}
In the following we describe the techniques we used to extend the current framework to accommodate our transformer based proposal and training procedure, we also describe the various experiments performed to compare methods. 

Natural Language Processing often deals with the problem of predicting one set of sequences from another set of sequences (for instance, a sequence of words from the sequence of acoustic features for automatic speech recognition tasks, or sequence of words in German from a sequence of words in English for language translation task, etc). As described in the above section, deep reinforcement learning also makes use of the sequence and we take inspiration from recent updates in NLP to propose a new method for DRL.
\begin{figure}[h]
    \centering
    \includegraphics[width=\linewidth]{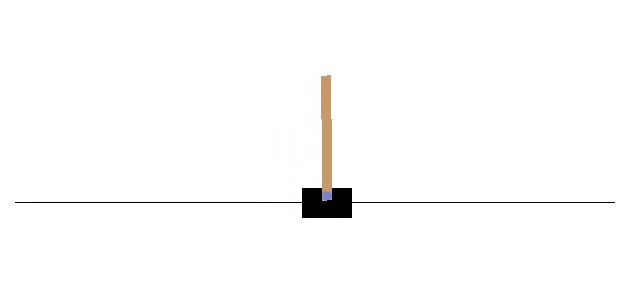}
    \caption{Frame from the Cartpole environment of OpenAI Gym. The task is to balance the pole on the cart, by moving the cart left or right}
    \label{fig:m1}
\end{figure}

\begin{figure*}
    \begin{subfigure}[b]{0.33\linewidth}
            \centering
            \includegraphics[width=0.5\linewidth]{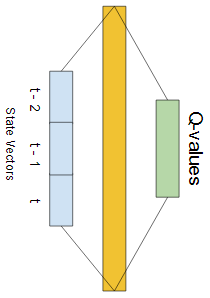}
            \caption{DQN}
            \label{fig:r1}
    \end{subfigure}%
    \begin{subfigure}[b]{0.33\linewidth}
            \centering
            \includegraphics[width=\linewidth]{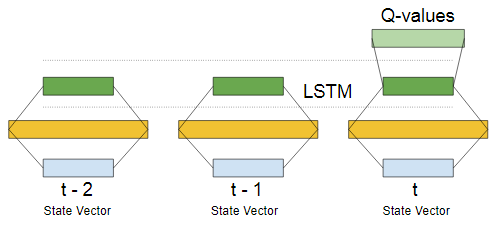}
            \caption{DRQN}
            \label{fig:r2}
    \end{subfigure}%
    \begin{subfigure}[b]{0.33\linewidth}
            \centering
            \includegraphics[width=0.8\linewidth]{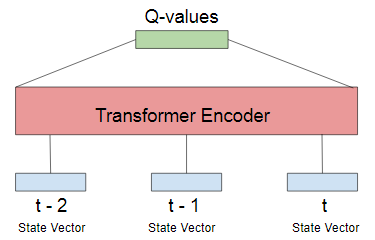}
            \caption{DTQN}
            \label{fig:r3}
    \end{subfigure}
    \caption{Different representative architectures. (a) DQN, (b) DRQN, (c) DTQN.}
    \label{fig:r4}
\end{figure*}

\subsection{Environment}
In our experiments, we set up an environment for the "Cartpole" game (using OpenAI gym \cite{openai}) where the goal is to balance the pole on the cart (i.e., prevent the pole from falling) by moving the cart left or right. Figure \ref{fig:m1} shows a frame from the game to visualize the environment. The set of actions $A$ consists of \{\textit{left}, \textit{right}\} and the environment provides a reward from the set of rewards $R$ consisting of \{$+1$, $-1$\}. For ever time-step where the pole does not fall the environment provides a reward of $+1$ and when the pole falls (i.e., the angle it makes from the cart crosses a certain threshold) the episode is completed and the agent receives a reward of $-1$. The typical deep reinforcement learning pipeline passes the frames (or sequence of frames) through a deep convolutional neural network to extract the useful features, which are further processed to estimate the \textit{value function ($V$)} or the \textit{state-action value function ($Q$)}. However, this results in model with relatively larger number of parameters, which also requires access to large GPUs to train the models using learning algorithms, not to mention that such algorithms take significantly longer to train. In order to overcome this problem we decided to use the RAM provided by the OpenAI environment describing the state of the game. In this case the state of the game can be described using the:
\begin{itemize}
    \item position of the cart (on the X-axis, from -4.8 to 4.8)
    \item velocity of the cart (from $-\infty$ to $\infty$)
    \item angle the pole makes with the cart (from -24 to 24)
    \item pole velocity at tip (from $-\infty$ to $\infty$)
\end{itemize}
However, to make the problem more interesting and difficult we set up a partially observable Markov decision process (POMDP), where the system dynamics are known to follow an MDP but the agent can not directly observe the states. In our experiments, the agent only observes the partial state consisting position of the cart and the angle pole makes with the cart. Different algorithms evaluated in this paper takes in the sequence of partially observed game states (a window of previous states preceding the current state) and predicts the action to be taken at the current state. The intuition is that the sequence of partially observed states should be sufficient for the algorithms to learn about the missing state features and act optimally.

In this work, we evaluate three different classes of algorithms namely, 1) Deep Q-Networks (DQN), 2) Deep Recurrent Q-Networks (DRQN) and 3) Deep Transformer Q-Network (DTQN). 

\subsection{DQN, DRQN, and DTQN}
Deep Q-Learning (DQN) uses a neural network to approximate the Q-value function. The state is given as the input and the Q-value of all possible actions is generated as the output. In a typical Deep Q learning setup, all the past experiences are first stored in the memory, the next action is then determined by using epsilon greedy policy with respect to current $Q$ values and the final loss is computed using equation \ref{e1}, where $S_t, a_t, r_t$ is the state, action taken and reward received at time-step $t$ and $S_{t+1}$ is the state at the next time-step.
\begin{align}
    L = ||r_t + \gamma \max_{a \in A}Q(S_{t+1},a) - Q(S_t, a_t)||_2^2
    \label{e1}
\end{align}
The term $r_t + \gamma \max_{a \in A}Q(S_{t+1},a)$ in equation \ref{e1} is known as the \textit{target} and will change erratically at every time-step as the Q values will change erratically at every time-step, in order to make the learning more stable we use a second copy of the deep neural network called \textit{target network}. The target network has the same architecture as the function approximator but with frozen parameters. For every $C$ iterations (a hyperparameter), the parameters from the prediction network are copied to the target network. This leads to more stable training. Figure \ref{fig:r1} shows a representative architecture for the DQN.
In our experiments, the input to DQN is the feature vector produced by concatenating the partially observed states from current and time-steps with the partially observed states of previous time-steps (4 time-steps in total, including current time-step).
\begin{figure}[H]
    \centering
    \includegraphics[width=0.6\linewidth]{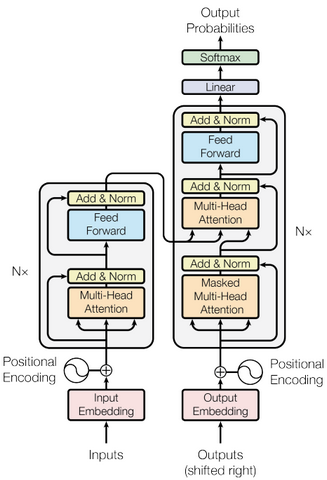}
    \caption{Transformer: encoder taking input sequence and decoder taking output sequences}
    \label{m2}
\end{figure}

Deep Recurrent Q-Learning (DRQN) uses a recurrent neural network to approximate the Q-value function. Sequence of states is given as the input and the network consists of RNN unit (LSTM/GRU), the output of the RNN unit at the final time-step is used to predict the Q-value of all possible actions is generated. Such networks are also often trained using the loss function defined equation \ref{e1}. Figure \ref{fig:r2} shows a representative architecture for DRQN.
In our experiments, the RNN unit (GRU) is fed the sequence of partially observed states, the output from the final time-step is used to predict the Q-value function.

The idea behind Deep Transformer Q-learning (DTQN) is to use a \textit{transformer} instead of RNN to extract the meaningful feature out of the input sequence. However, transformers were designed to work for sequence to sequence (seq2seq) tasks such as automatic speech recognition or language translation. But in this work we do not have s sequence to sequence task, rather we need to predict the $Q$ value function given the input sequence.
\begin{figure}[H]
    \centering
    \includegraphics[width=0.4\linewidth]{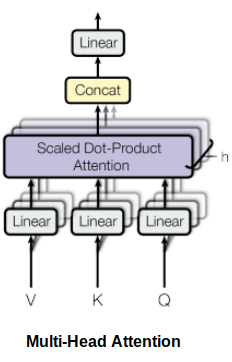}
    \caption{Transformer: encoder taking input sequence and decoder taking output sequences}
    \label{m3}
\end{figure}

\begin{figure*}
\includegraphics[width=0.5\linewidth]{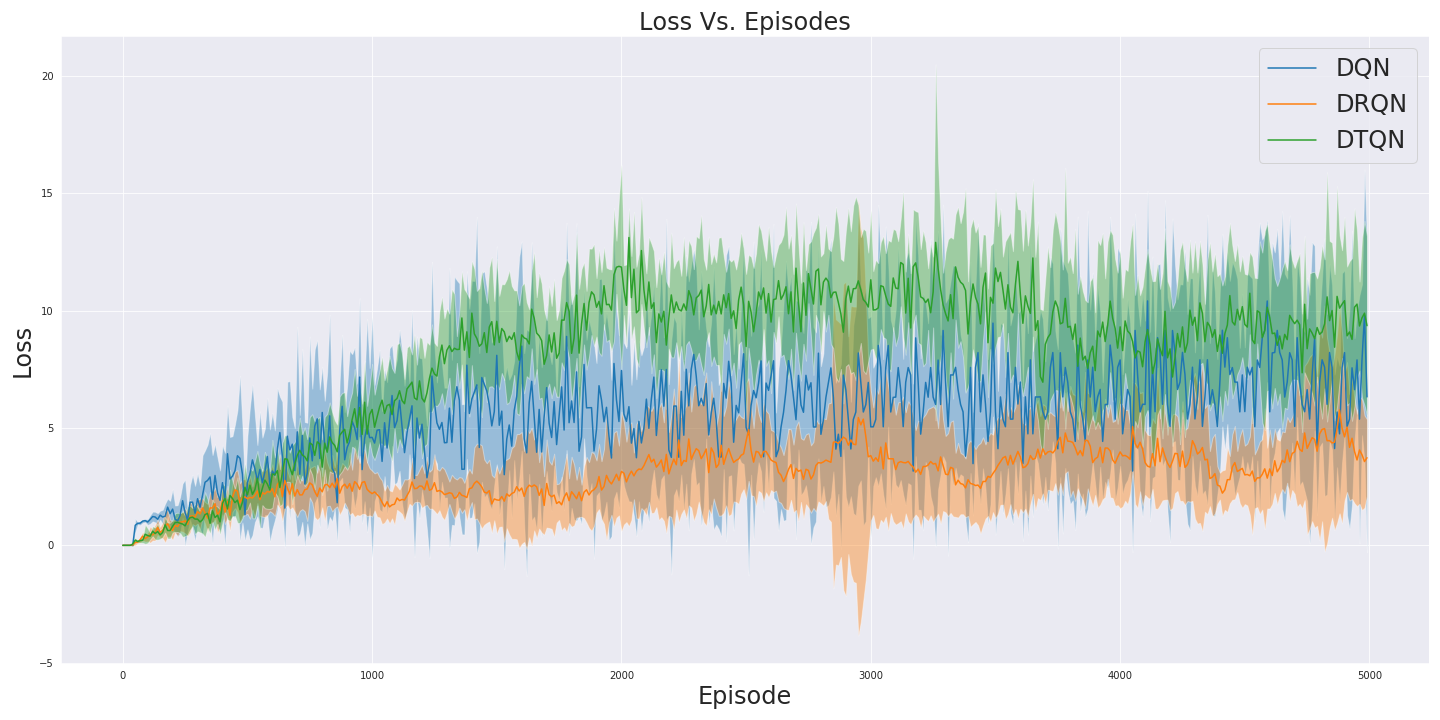}
\includegraphics[width=0.5\linewidth]{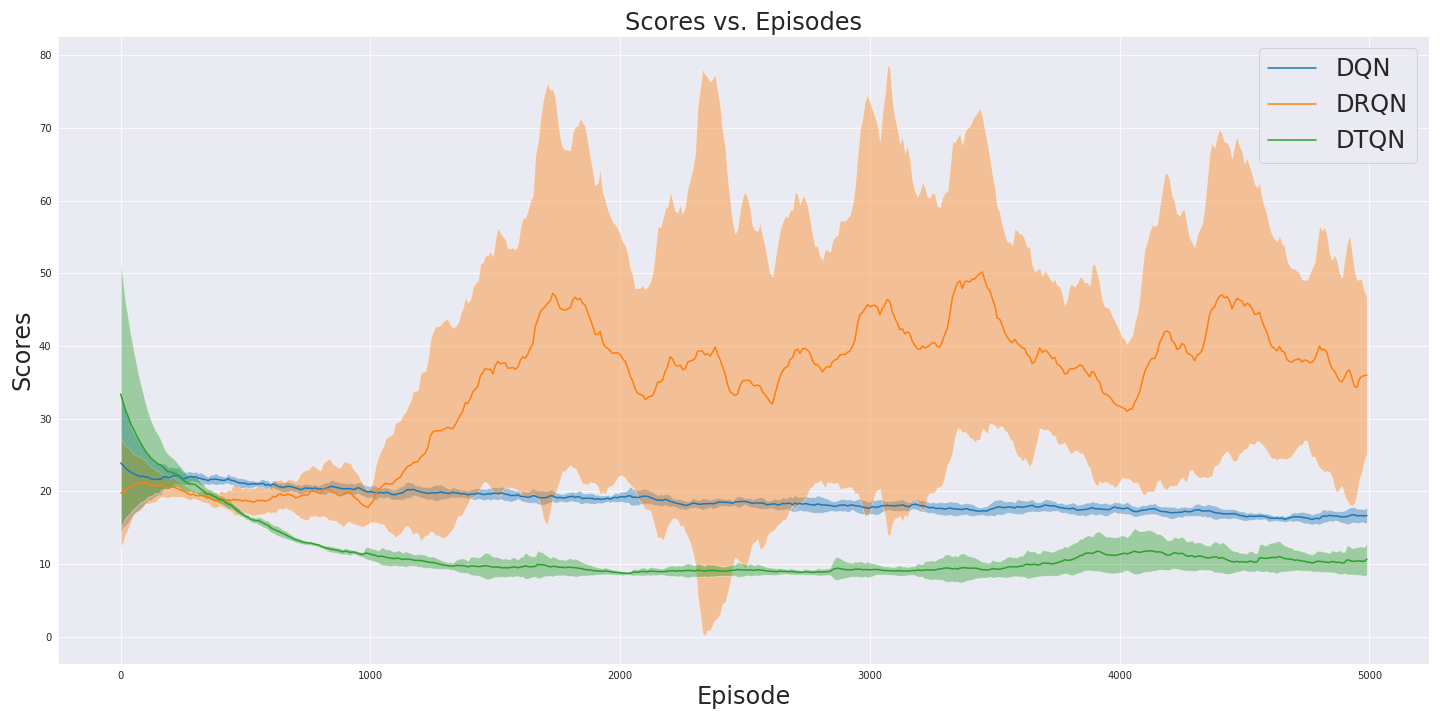}
\end{figure*}

\begin{figure}
    \centering
    \begin{subfigure}[b]{\linewidth}
            \centering
            \includegraphics[width=\linewidth]{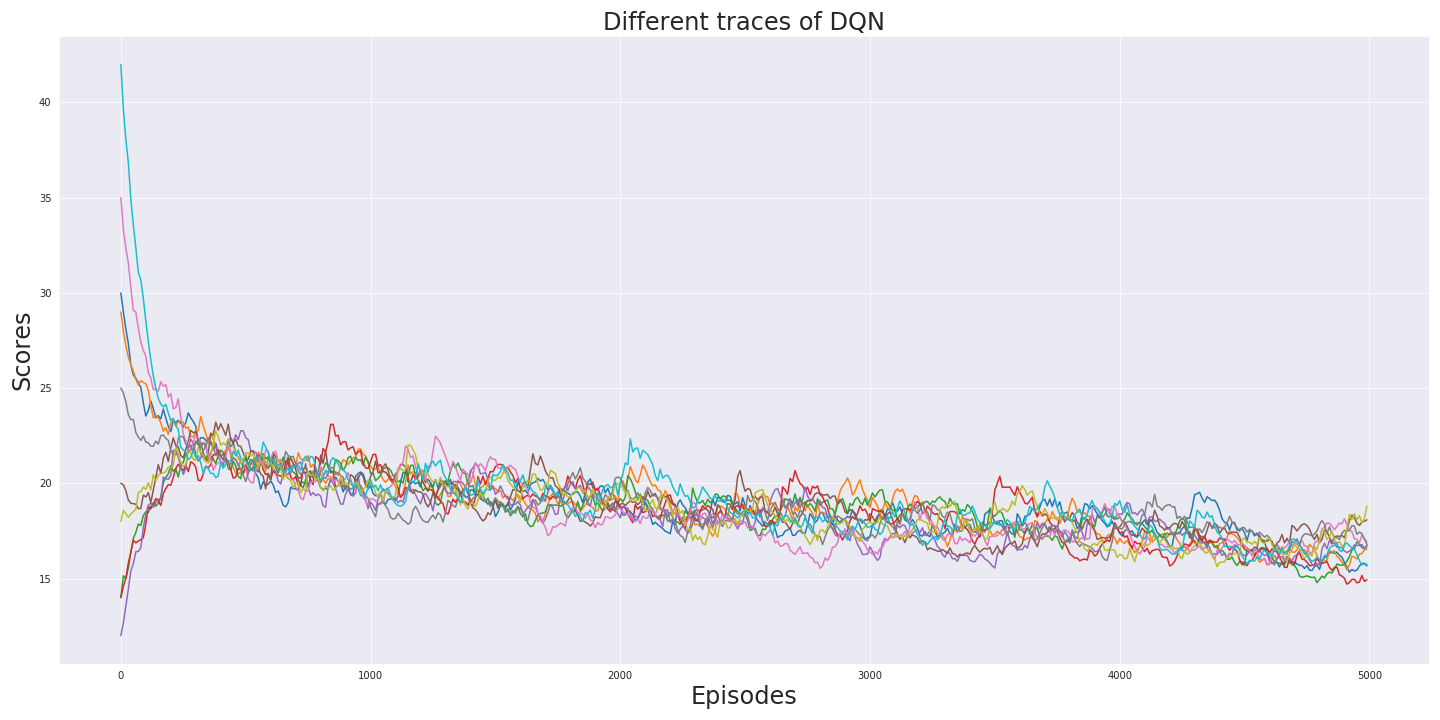}
            \caption{traces of DQN}
            \label{fig:d1}
    \end{subfigure}%

    \begin{subfigure}[b]{\linewidth}
            \centering
            \includegraphics[width=\linewidth]{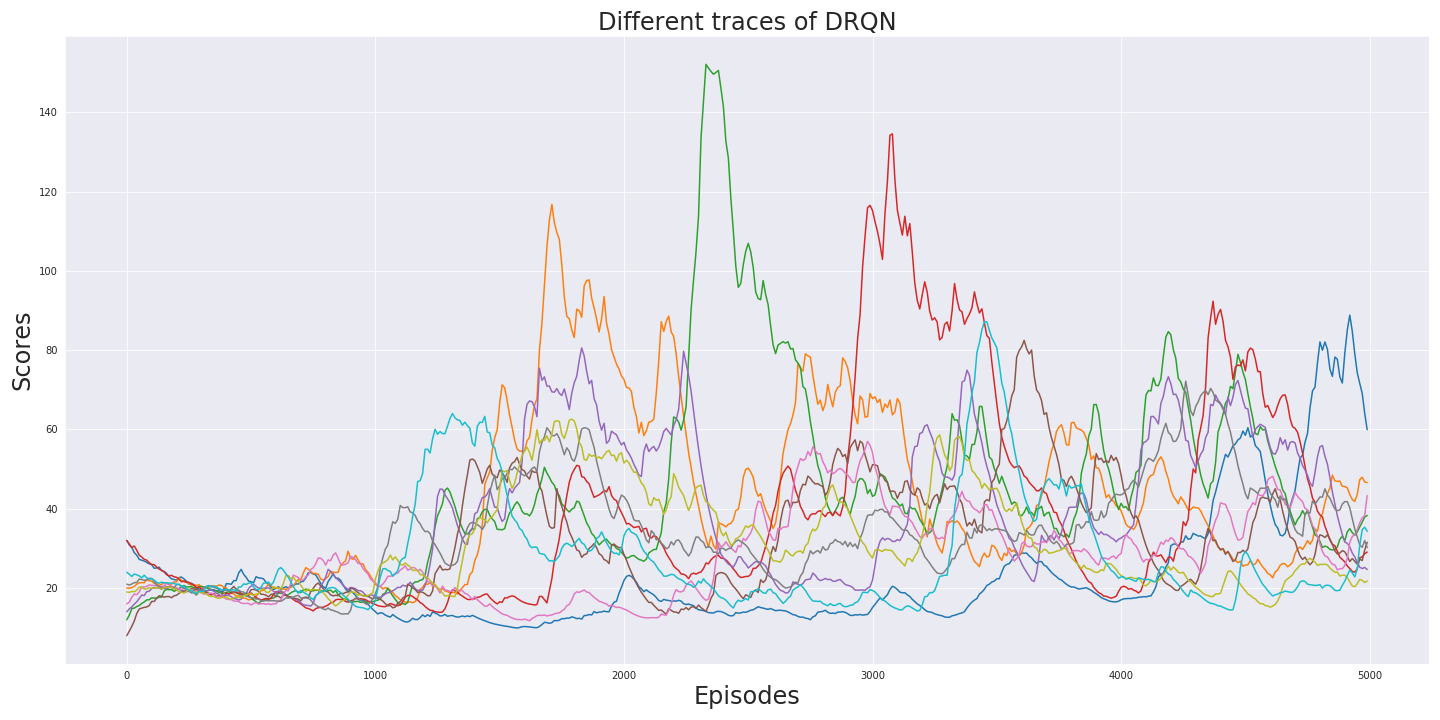}
            \caption{traces of DRQN}
            \label{fig:d2}
    \end{subfigure}%

    \begin{subfigure}[b]{\linewidth}
            \centering
            \includegraphics[width=\linewidth]{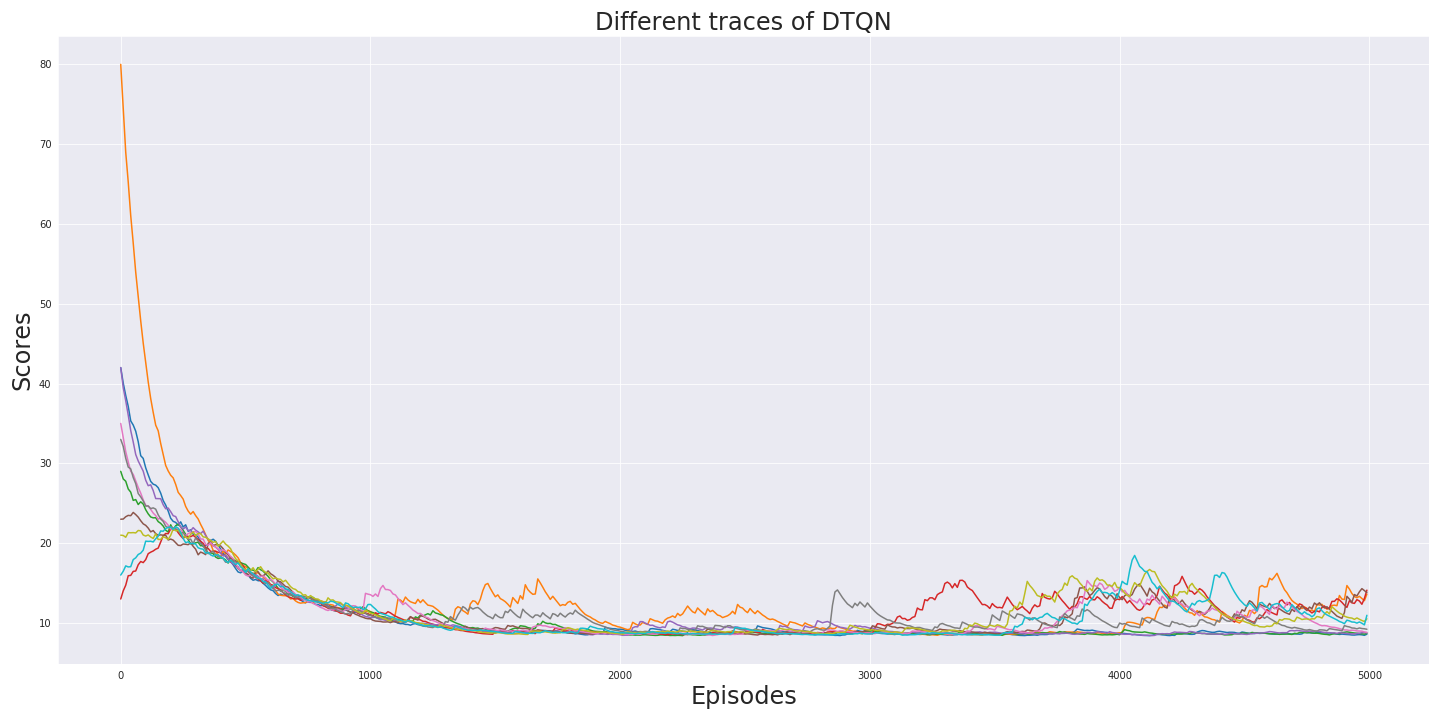}
            \caption{traces of DTQN}
            \label{fig:d3}
    \end{subfigure}
    \caption{Scores vs Episodes for multiple runs of different algorithms. (a) DQN, (b) DRQN, (c) DTQN.}
    \label{fig:r4}
\end{figure}

The transformer model consists of encoder and decoder modules, as shown in figure \ref{m2}. The encoder module processes the input sequence of embeddings (tokens from input sequence are first embedded and then passed through the network) through a multi-head attention module followed by a feed-forward neural network. In our experiments we only use the encoder module (since our task is not Seq2Seq) to extract the features from input sequence which are further used to predict the Q-values. The multi-head attention module in the encoder refers to the self-attention layer which is a mechanism relating different positions of a single sequence in order to compute a representation of the sequence. Figure \ref{m3} shows the multi-head attention module used in the encoder, $V$, $K$, $Q$ denotes the value vector, key vector, query vector. Figure \ref{fig:r3} shows a representative architecture for DTQN.

\section{Results and Discussions}
\label{sec:results}

We did multiple experiments for each of the following DQN, DRQN and DTQN based reinforcement learning algorithms. Each algorithm was trained for 5000 episodes and we ran 10 different instances for each of the algorithms with random initialization. Figure \ref{fig:r4} shows the value of the scores over episodes for different runs. We see that DQN and DTQN performed worse compared to our DRQN model. While on average the scores for the DQN and DTQN are decreasing, we see that there are a few experiments where the scores improve during training as shown in the graph, but for a majority of the traces, the scores did not improve when using DQN or DTQN. However, the results show significant improvement in scores when using the DRQN model.

As we used location and angle as the only inputs to our neural networks determine the next action, the scores and loss function did not show very promising trends in all three cases. Still, it was possible to make a distinction as to which one of the three: DQN, DRQN, or DTQN performed better by analyzing their relative scores and loss values. DRQN gave the best score results with the maximum reaching to 135 in one of the test cases. DTQN performed the worst on average. 

We believe that a transformer-based approach is indeed not suitable for solving the reinforcement learning problem at least in cases where input is taken in the form of a pair of location value and angle value. Reason for the failure can be pointed out as follows: LSTM based approach (DRQN) captures the temporal attributes of the sequence of inputs whereas transformer-based approach tries to put attention on different time-steps of the sequence to obtain the representation for input sequence and does not explicitly tries to capture the temporal aspect of the sequence. As our problem is more of temporal related as the next state is the state at the next time step, LSTM achieves better results. We would also like to add that during our course of work we found that training RL algorithm, for video games, based on DQN or neural networks, in general, is difficult to train and the performance greatly depends upon the random initialization. As can be seen from the traces, some of the random initialization perform too bad while a few perform well.

\section{Conclusions and Future Work}
In this work, we propose a new transformer based model for reinforcement learning, the inspiration for the same is derived from the fact that the recent advancements in NLP have been achieved by moving away from RNNs and introducing the new transformer based model. Even though the standard transformer consists of both encoder and decoder modules as it's designed to perform Seq2Seq task, we decided to use only the encoder module with the multi-head attention mechanism from the transformer to extract important features from the states to learn the Q-values. We perform experiments on a POMDP and compared multiple algorithms. We conclude that using the state representation provided by the RAM of the game (X-coordinate of the cart and the angle made by pole with the cart), the conventional DRQN (based on RNNs such as LSTM and GRU) perform better than DQN or the proposed DTQN. As part of future work, we intend to compare different algorithms using different state representation like images instead of RAM values.

\bibliographystyle{IEEEbib}
\bibliography{refs}

\end{document}